\RequirePackage{fix-cm}
\documentclass[twocolumn]{svjour3}          
\smartqed  
\usepackage{graphicx}
\usepackage{times}
\usepackage{epsfig}
\usepackage{amsmath}
\usepackage{amssymb}
\usepackage{algorithm}
\usepackage{algorithmicx}
\usepackage{algpseudocode}
\usepackage{booktabs}

\journalname{Machine Vision and Applications}
\begin{document}
\title{Bimodal Stereo: Joint Shape and Pose Estimation from Color-Depth Image Pair\thanks{The research is supported by National Natural Science Foundation of China (NSFC) under Grant No.91520301.}
}


\author{Chi Zhang$^1$         \and
        Yuehu Liu$^1$         \and
        Ying Wu$^2$           \and
        Qilin Zhang$^3$       \and
        Le Wang$^1$        
}


%

\institute{$^1$Institute of Artificial Intelligence and Robotics, Xi'an Jiaotong University, No.28 Xianning West Road, Xi'an, Shaanxi, China\\
%
$^2$Department of Electronical Engineering and Computer Science, Northwestern University, 2145 Sheridan Road, Evanston, Illinois, 60208, USA\\
$^3$HERE Technologies, 425 W Randolph St, Chicago, Illinois, 60606, USA\\
}

\date{Received: date / Accepted: date}

\maketitle

\begin{abstract}
Mutual calibration between color and depth cameras is a challenging topic in multi-modal data registration. In this paper, we are confronted with a "Bimodal Stereo" problem, which aims to solve camera pose from a pair of an uncalibrated color image and a depth map from different views automatically. To address this problem, an iterative Shape-from-Shading (SfS) based framework is proposed to estimate shape and pose simultaneously. In the pipeline, the estimated shape is refined by the shape prior from the given depth map under the estimated pose. Meanwhile, the estimated pose is improved by the registration of estimated shape and shape from given depth map. We also introduce a shading based refinement in the pipeline to address noisy depth map with holes. Extensive experiments showed that through our method, both the depth map, the recovered shape as well as its pose can be desirably refined and recovered.
\end{abstract}

\section{Introduction}
\label{intro}
With the development of depth sensors (time-of-flight (ToF) camera, laser range scanner, structured light scanner, etc.), researchers have been dedicating themselves on utilizing both color and depth information for 3D scene reconstruction and related topics. The calibration of color and depth camera\cite{HanTOC13}, especially extrinsic calibration (i.e. pose estimation between cameras), thus becomes a prerequisite in building such applications. Existing manual and semi-automatic methods\cite{HerreraPAMI12,ZhangICME11} are solving this problem in the settings where multiple color images and checkerboard are required.\par
\begin{figure}
\centering
       \includegraphics[width=0.5\textwidth]{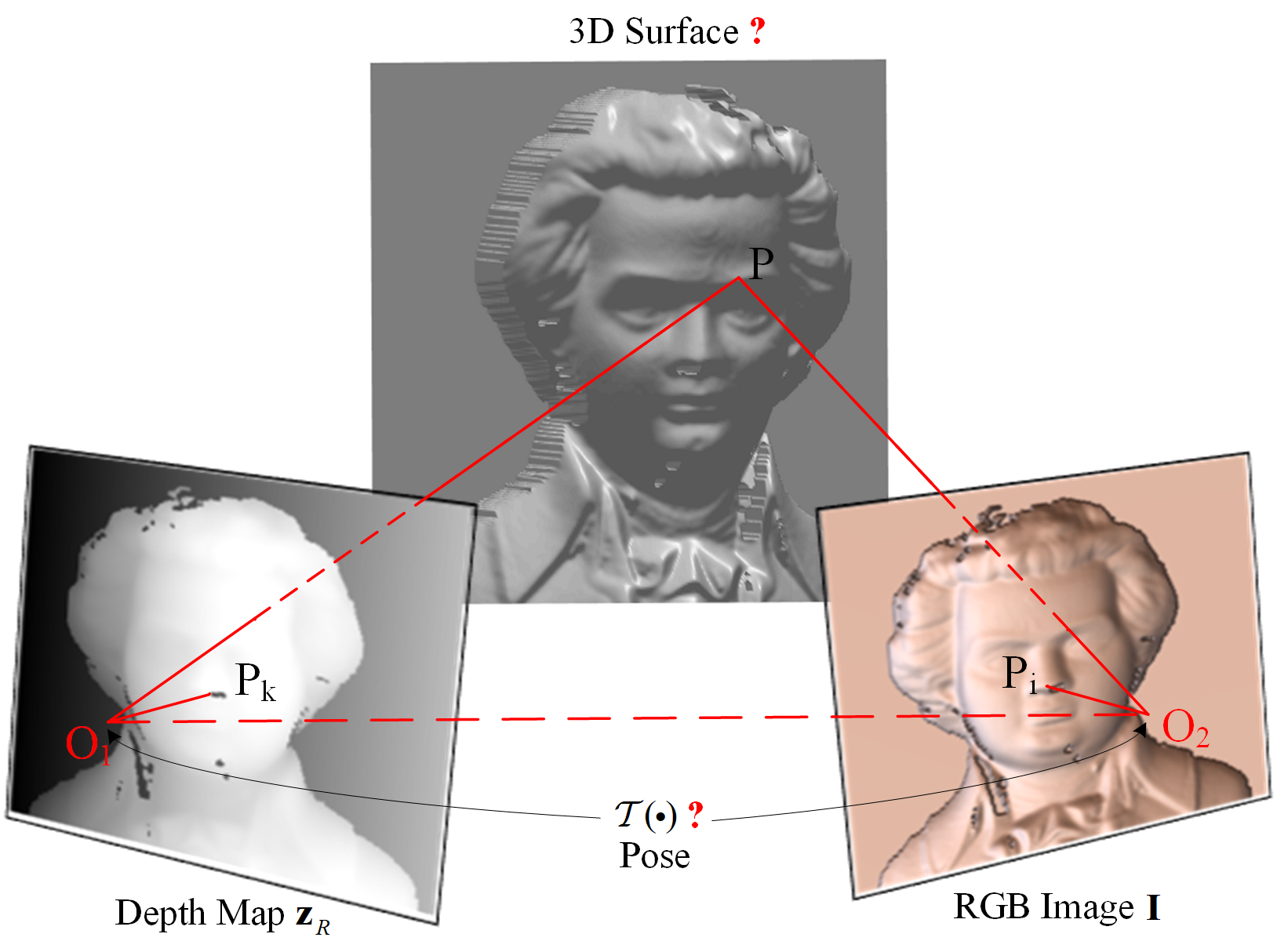}
      \caption{Problem Settings of Bimodal Stereo. In this paper, the pose $\mathcal{T}(\cdot)$ between a depth map $\mathbf{z}_R$ and a color image $\mathbf{I}$ is to be solved.   }
\end{figure}\label{fig1}
In this paper, we are confronted with a similar but different problem, which aims to solve camera pose from a pair of an uncalibrated color image and a depth map from different views automatically (Illustrated in Figure~\ref{fig1}). Such problem has varieties of difficulties that existing method may be failed. Firstly, 3D coordinates of feature points in image coordinate system could not be easily obtained from an uncalibrated color image. Moreover, point-wise correspondence may be difficult to generate between this pair of multi-modal images.\par 
\begin{figure*}[!htp]
\centering
       \includegraphics[width=\textwidth]{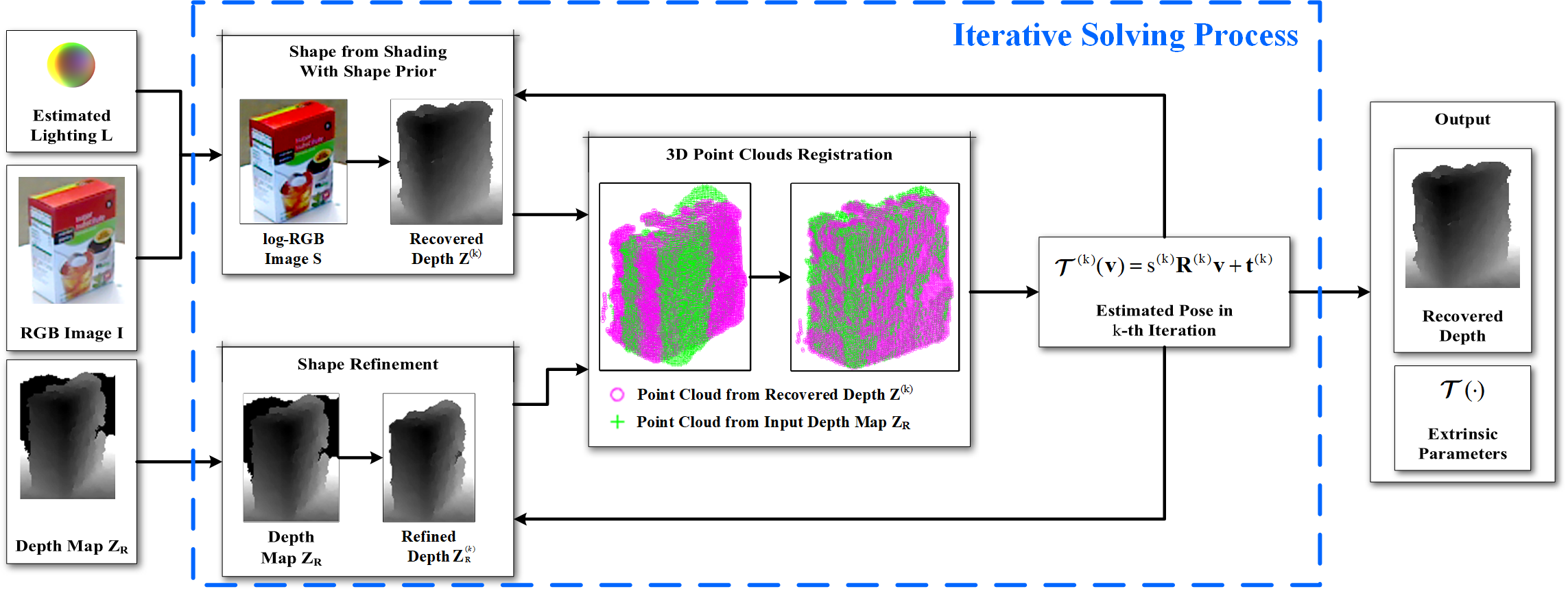}
      \caption{The proposed iterative joint shape and pose estimation framework. In the $k$-th iteration, \textbf{Shape-from-Shading} $\ell$: Given an input RGB image $\mathbf{I}$, per-pixel depth $\mathbf{z}^{(k)}$ and normal map $\mathbf{n}^{(k)}$ is estimated according to the pre-obtained lighting parameter. \textbf{3D Point Clouds Registration} $f$: the pose $\mathcal{T}^{(k)}(\cdot)$ is updated between estimated depth $\mathbf{z}^{(k)}$ and input depth $\mathbf{z}_R$. \textbf{Shape Refinement} $g$: the input depth $\mathbf{z}_R$ is refined based on the newly shape estimation $\mathbf{n}^{(k)}$ and pose $\mathcal{T}^{(k)}(\cdot)$. Then the method continues into the new stage of estimation iteratively, with $\mathcal{T}^{(k)}(\mathbf{n}_R)$, the product of current pose estimation $\mathcal{T}^{(k)}(\cdot)$ and the input normal map $\mathbf{n}_R$, serving as the shape prior in the new SfS procedure. }
      \label{fig2}
\end{figure*}
To solve such ``Bimodal Stereo" problem, we take the assumption of Spherical Harmonics (SH) illumination \cite{RamamoorthiSIGGRAPH01} based image formation process and propose an iterative Shape-from-Shading (SfS) based framework to estimate shape and pose simultaneously, shown in Figure~\ref{fig2}. In each iteration, the input data is processed through a pipeline of three procedures, namely Shape from shading, Point clouds registration and Shading based shape refinement. This method is based on the notation that given sufficient shape prior the SfS reconstruction could be largely improved (see details in Section \ref{prior}). In further, more accurate pose can be estimated through registration between reasonable reconstruction and given depth map , which in return benefits the SfS process. Therefore, given the estimated lighting parameters, an uncalibrated color image and depth map, the proposed EM-like algorithm is able to obtain plausible shape and pose. \par
The contribution of our work is listed as follows:\par
Firstly, an interesting problem of Bimodal Stereo is introduced. This problem is non-trivial since it aims to automatically solve the pose between a pair of a single uncalibrated image and a depth map. \par
Besides, Spherical Harmonics Shape from shading method is utilized to solve the Kinect extrinsic calibration fully automatically and targetlessly. To our knowledge, it is the first attempt to register multi-modal images by image formation model, other than feature-based methods \cite{HeinrichMIA11} or mutual information (MI) based methods\cite{PluimMICCAI00}. With Spherical Harmonics modelling the illumination, our method becomes a well-posed one, which can be effectively solved by non-linear least square methods.\par
Moreover, the cost function of SH Shape from shading is regularized by imposing a simple shape prior term, which bridges the shading information from color image and the shape information from depth map by the estimated pose between two cameras. \par
This paper will proceed as follows: In Section \ref{2}, related works are reviewed and compared with the proposed method. In Section \ref{3} our method will be formulated as a minimization problem over Shape from Shading, point clouds registration and shading based shape refinement. In Sections \ref{4}, \ref{5}, and \ref{6} the cost function and solution of each process are described respectively. In Section \ref{7} we will not only show the results on synthesized data and real data, but also conduct some quantitative experiments to validate the assumption and the robustness of our method. In the last section, this paper will be concluded while the limitations of proposed method are discussed.
\section{Related Works} \label{2}
Although the physical process of image formation is still under discussion yet, the majority of researchers in computer vision area would agree that images convey shape information implicitly. In 1975 Horn proposed the well-known ill-posed inverse problem called "Shape from Shading (SfS)" ,which aims to recover shape from a single grayscale image. The problem is based upon the assumption that the shading information of an image is formed by a known light source and surface normals when the albedo of the surface is constant.  \par
From then on, many efforts have been made to solve this problem under simplified assumptions on illumination. The most classical setting is to recover the shape of a Lambertian surface, i.e. a diffusely reflecting surface, under the lighting of a distant point source. From early 80's to now, various method proposed, including directly numerical solutions\cite{IkeuchiAI81,PradosICCV03},linear approximation\cite{TsaiIVC94} and minimization\cite{CrouzilTPAMI03},achieved success in solving the problem under this oversimplified assumption. Since Basri \textit{et al.}\cite{BasriTPAMI03} have proved images of a convex Lambertian object under a wide variety of lighting conditions can be approximated accurately by a 9D SH coefficients vector, researcher starts to pay attention on SH illumination model\cite{RamamoorthiSIGGRAPH01} which achieved great success in high quality rendering. Instead of solving the surface gradients or depth in existing method, our method seek for the unit surface normal which satisfies the RGB-shading model, thus the problem become well-posed if the three coefficients vectors for R, G and B channel respectively are whitened\cite{BarronPAMI15}. Moreover, we posed simple shape prior term in our SfS method as a bridge between the two multi-modal images instead of other various regularization terms\cite{ZhangTPAMI99}, which makes our cost function parsimony while still strong enough.  \par
Currently, while some of the researchers tried to add some constraints making SfS problem well-posed\cite{PradosCVPR05}, others \cite{YuCVPR13,WuTOG14}turned to embed SfS techniques to refine the noisy and holey depth map obtained by Kinect. In Yu's method\cite{YuCVPR13} the SH coefficients vectors for each channel are not whitened hence cannot ensure the property well-posed and generally slower with more terms than ours. Wu's method \cite{WuTOG14} is the first real-time shading based shape refinement method. However, since the method directly refines the depth information, it is hard to impose our regularization of shape prior. Our method, instead, can be viewed as a symmetric version of proposed SfS method. Therefore it is also a well-posed problem which solves surface normal at each pixel.
\par
As stated before, the proposed Bimodal stereo problem can be viewed as an automatically extrinsic calibration between color camera and depth camera of a Kinect. Though factory calibrated, the calibration of Kinect is still important to applications in high-accuracy 3d reconstruction and adding new cameras. Generally, existing methods \cite{HerreraPAMI12,ZhangICME11} are derived from Zhang's RGB camera calibration method\cite{ZZhangTPAMI00}. Both methods requires calibration card to calibrate the color image first before extrinsic calibration by feature matching on both images. Our method, however, does not require RGB camera calibration. Moreover, the pose between cameras is solved automatically, which meets the requirement of Bimodal Stereo problem.  
\section{Problem Formulation}\label{3}
\subsection{Method Overview}
In this paper an iterative computational framework is proposed to solve problem Eq.~\eqref{eq13}, as shown in Algorithm~\ref{alg1}. and illustrated in Figure~\ref{fig2}. The input of the system is depth map $z_R$ and RGB log-shading $\{S_{ij}\}_{i=1,...,M,j=1,2,3}$, where $M$ is the total amount of pixels and $j=1,2,3$ denotes the R, G or B log-values of the i-th pixel. The goal is to estimate normal $\mathbf{n}^{(k)}$ and depth $\mathbf{z}^{(k)}$ from the shading information and the refined depth $z_R^{(k)}$ as well as the motion $\mathcal{T}^{(k)}$ between RGB camera and depth camera. In the loop, we first estimate $\mathbf{n}$ in a shape-from-shading way based on Spherical Harmonics Illumination model. Then the depth $\mathbf{z}^{(k)}$ can be obtained by solving the partial differential equations of estimated surface gradients $p,q$ from $\mathbf{n}^{(k)}$. Afterwards, the transformation $\mathcal{T}^{(k)}(\mathbf{P})=s\mathbf{RP}+\mathbf{t}$ is estimated by registering the point clouds $P_c^{(k)}=\{(x,y,z^{(k)})\}$ from $z^{(k)}$and $P_c$ from $z_R^{(k)}$. Also, a mapping describes the point wise correspondence between the two point clouds is established during the registration. The last procedure of one iteration is to refine the overlapping area of given depth $z_R$ and its normal $n_R$ based on the rotation $\mathbf{R}$ and shading $\{S\}$ in this area. Then the improved normal $\mathbf{n}_R^{(k)}$ is utilized as the initial input in the next iteration. The loop will end when the norm of the difference two matrix $\mathbf{R}^{(k)}$ and $\mathbf{R}^{(k-1)}$ is less than a threshold set by user.
\subsection{Formulation}
We call our problem formulation for iteratively estimating shape information from the unknown correspondence of a RGB image and a Kinect depth map from different views as "Bimodal Stereo". Bimodal Stereo can be thought of as a shape completion and refinement by multi modal data registration. It is an integrated iterative process including three stages Shape from Shading(SfS), Point Cloud Registration and Shape refining. The "Bimodal Stereo" problem formulation is :
\begin{equation} \label{eq13}
\underset{\mathbf{n}^{(k)},\mathbf{z}^{(k)},\mathcal{T}^{(k)}(\cdot)}{\mathop{\min}}\,\quad \ell (\mathbf{n}^{(k)},\mathbf{z}^{(k)},\mathcal{T}^{(k)}(\cdot))+ f(\mathbf{z}^{(k)},\mathcal{T}^{(k)}(\cdot))+g(\mathbf{n}^{(k)},\mathcal{T}^{(k)}(\cdot)),
\end{equation}
where $\mathbf{n}^{(k)}$ and $\mathbf{z}^{(k)}$ is the estimated normal and depth map of the surface from the RGB shading information. $\mathcal{T}(\mathbf{P})=s\mathbf{RP}+\mathbf{t}$ is the estimated 3D transformation for a point cloud $\mathbf{P}$. In transformation $\mathcal{T}$, $s$ is the scale factor, $\mathbf{R}$ is the $3\times 3$ rotation matrix between point clouds $P_c^{(k)}=\{(x,y,z^{(k)})\}$ and $P_c$ from depth map and $\mathbf{t}$ is the $3\times 1$ translation vector. These parameters are estimated by minimizing the sum of cost functions $\ell (\mathbf{n}^{(k)},\mathbf{z}^{(k)},\mathcal{T}(\cdot))$,$f(\mathbf{z}^{(k)},\mathcal{T}^{(k)}(\cdot))$ and $g(\mathbf{n}^{(k)},\mathcal{T}^{(k)}(\cdot))$ for SfS, registration and shape refining respectively.

Therefore the detailed description of our method is divided into 3 parts. SH based Shape from Shading is explained in the next chapter while registration and shape refinement are discussed in Chapter 4 and 5 respectively. In Chapter 6 we'll show the validation and quantitative analysis of our method.
\begin{algorithm} 
\caption{The framework of Bimodal Stereo} 
\label{alg1}
\begin{algorithmic}[1] 
\Require $\{S_{ij}\}$, $\mathbf{z}_R$, $k = 1$
\Ensure $\mathbf{n}^*,\mathbf{z}^*,\mathcal{T}^*(\cdot)=\{s,\mathbf{R,t}\},\mathbf{z}_R^*$
\State $R^{(0)} \gets \mathbf{1}$, where $\mathbf{1}$ is $3\times 3$ identity matrix;
\State $R^{(1)} \gets \mathbf{J}$, where $\mathbf{J}$ is a $3\times 3$ matrix whose elements are all larger than 1;
\While {Norm ($R^{(k)}-R^{(k-1)}>$ Threshold)}
\State $R^{(k)} \gets R^{k-1}$
\State $\mathbf{n}^{(k)} \gets$ ShapefromShading($\{S_{ij}\},\mathbf{R}^{k},\mathbf{z}_R$)
\State $\mathbf{z}^{(k)} \gets$ ShapefromSurfaceGradient($\mathbf{n}^{(k)}$)
\State $\mathbf{P}_c^{(k)} \gets$ Depth2PointCloud($\mathbf{z}^{(k)}$)
\State $s^{(k)},\mathbf{R}^{(k)},\mathbf{t}^{(k)} \gets$ PointCloudRegistration($\mathbf{P}_c^{(k)},\mathbf{P}_c$)
\State $z_R^{(k)} \gets$ ShadingbasedRefinement($\{S_{ij}\},\mathbf{n}^{(k)},\mathbf{R}^{(k)},\mathbf{z}_R$) 
\EndWhile 
\end{algorithmic}
\end{algorithm}

\section{Spherical Harmonics Shape from Shading}\label{4}
In this section, we present our Shape from Shading method. Based on the assumption that the RGB image is formed by Spherical Harmonics Lighting model and surface normals, the SfS process can be formulated as an inverse problem of this image formation model. By posing the shape prior in the SfS formulation, the correlation between estimated shape and estimated pose is set up. Such problem is well-posed and can be solved by nonlinear least square method. Once obtained the solution normal, the 3D point cloud of estimated shape can be computed through the least squares surface reconstruction.
\subsection{Spherical Harmonics Lighting Model}
Traditional shape from shading methods assume the illumination model as a sufficiently remote point source. Such simple model though effective in shape reconstruction from synthetic images, could not perform well with images acquired in the natural world. The Spherical Harmonics (SH) Illumination however provide a more accurate while not complex description of the natural lighting. 
Instead of modelling the illuminant in whole, the SH model assumes that at an arbitrary point in the 3D space, the light direction could be anisotropy. Such complex lighting condition could be parametrized by a 27 (color, 9 dimensions per RGB channel) dimensional vector, $L$.\par
After fitting a multivariate Gaussian to the SH illuminations in the training set, the estimated SH coefficients vector $L$ is obtained by minimizing the following cost function\cite{BarronPAMI15}
\begin{equation}
h(\mathbf{L})=\lambda_L(\mathbf{L}-\mu_L)^{T}\mathbf{\Sigma}_L^{-1}(\mathbf{L}-\mu_L),
\end{equation}
where $\mu_L$ and $\Sigma_L$ are learned parameters of the Gaussian, while $\lambda_L$ is the multiplier learned on dataset. To ensure the coefficients vector of each channel to be independent with each other, we preprocess these vectors by whitening transformation described in \cite{BarronPAMI15}.\par 
The log-intensity at pixel $i$, $\log S_i$ of a Lambertian surface under SH illumination could be described in the following equation:
\begin{equation}\label{eq11}
\log S_i={\mathbf{\hat{n}}_i}^T \mathbf{M_j} \mathbf{\hat{n}}_i
\end{equation}
where $\mathbf{\hat{n}}_i$ is a 4-dimensional vector:
\begin{equation}
\left[
\begin{matrix}
-\frac{{{p}_{i}}}{\sqrt{1+p _{i}^{2}+q _{i}^{2}}}, & -\frac{{{q }_{i}}}{\sqrt{1+p _{i}^{2}+q _{i}^{2}}}, & \frac{1}{\sqrt{1+p _{i}^{2}+q _{i}^{2}}}, & 1  
\end{matrix}
\right]^T.
\end{equation}
In this Equation, $p_i=\frac{\partial z_i}{\partial x}$ and $q_i=\frac{\partial z_i}{\partial y}$ is the partial derivative of the depth map $z$ at pixel $i$ along $x$ and $y$ axis, respectively. It is worth to point that $\mathbf{\hat{n}_i}=\left[ \begin{matrix}
\mathbf{n}_i^T, & 1
\end{matrix} \right]^T$
where $\mathbf{n}_i^T$ is the surface normal at pixel $i$. \\
$\mathbf{M}_j$ is a $4\times 4$ matrix for the j-th RGB channel, described by
\begin{equation}
   \mathbf{M}_j=\left[ \begin{matrix}
   {{c}_{1}}{{L}_{j,9}} & {{c}_{1}}{{L}_{j,5}} & {{c}_{1}}{{L}_{j,8}} & {{c}_{2}}{{L}_{j,4}}  \\
   {{c}_{1}}{{L}_{j,5}} & -{{c}_{1}}{{L}_{j,9}} & {{c}_{1}}{{L}_{j,6}} & {{c}_{2}}{{L}_{j,2}}  \\
   {{c}_{1}}{{L}_{j,8}} & {{c}_{1}}{{L}_{j,6}} & {{c}_{3}}{{L}_{j,7}} & {{c}_{2}}{{L}_{j,3}}  \\
   {{c}_{2}}{{L}_{j,4}} & {{c}_{2}}{{L}_{j,2}} & {{c}_{2}}{{L}_{j,3}} & {{c}_{4}}{{L}_{j,1}}-{{c}_{5}}{{L}_{j,7}}  \\
\end{matrix} \right] 
\end{equation}
where 
\[ \begin{aligned}
&{{c}_{1}}=0.429043,{{c}_{2}}=0.511664,{{c}_{3}}=0.743125,\\
&{{c}_{4}}=0.886227,{{c}_{5}}=0.247708 
\end{aligned} \]
and $\mathbf{L}_j$ denote the SH coefficients of j-th RGB channel.
 
\subsection{Shape from Shading with shape prior}
Recovering surface normal $\mathbf{n}_i^*$ from the corresponding brightness $ S_{ij}, j=1,2,3$ could be viewed as a constrained optimization problem for $\mathbf{n}_i^*$ on the following equation:
\begin{equation} \label{eq1}
\begin{aligned}
\underset{\mathbf{n}_i^*}\min \: \ell (\mathbf{n}_i^*,z_i^*,\mathcal{T}(\cdot)) &=\ell_1(\mathbf{n}_i^*)+{\lambda }_{2}\ell_2(\mathbf{n}_i^*,\mathbf{n}_{R_h},\mathcal{T}(\cdot)))\\
&\mathbf{s.t.}\;  {\mathbf{n}_i^*}^T\mathbf{n}_i^*=1
\end{aligned}
\end{equation}
where $j=1,2,3$ denotes the R,G,B channel and the corresponding point of the $i$-th pixel is $h$-th point on the given depth map. $\mathcal{T}(\cdot)$ is the estimated pose. The constraint is the norm regularization which enforces normals to be of unit length.
We can pose the constraint ${\mathbf{n}_i^*}^T\mathbf{n}_i^*=1$ into $\ell (\mathbf{n}_i^*,z_i^*,\mathcal{T}(\cdot))$ so that the solution of Eq.~\eqref{eq1} can be converted to the following unconstrained optimization problem:
\begin{equation}
\begin{aligned}
\underset{\mathbf{n}_i^*}\min \: \ell (\mathbf{n}_i^*,z_i^*,\mathcal{T}(\cdot)) & =\ell_1(\mathbf{n}_i^*)+{\lambda }_{2}\ell_2(\mathbf{n}_i^*,\mathbf{n}_{R_h},\mathcal{T}(\cdot))\\
& + \lambda_3({\mathbf{n}_i^*}^T\mathbf{n}_i^*-1)
\end{aligned}
\end{equation}
In Eq.~\eqref{eq1}, ${{\ell }_{1}}(\mathbf{n}_i^*)$ is the \textit{Brightness Constraint} measuring the total squared brightness error over the grayscale image $S$:
\begin{equation} \label{eq12}
{{\ell }_{\text{1}}}(\mathbf{n}_i^*)={{(\log S_{ij}-{\mathbf{\hat{n}}_i}^T \mathbf{M}_j \mathbf{\hat{n}}_i )}^{2}},
\end{equation}
The second term ${{\ell }_{2}}(\mathbf{n}_i^*,\mathbf{n}_{R_h},\mathcal{T}(\cdot)))$ serves as \textit{Shape Prior}, which enforces the estimated normals $\mathbf{n}_i^*$ of $z_i^*$ to be close to the rotated ground truth normals $\mathbf{n}_{R_h}$ from depth map $z_{R_h}$. This term is modelled as
\begin{equation}\label{eq10}
{\ell }_{2}(\mathbf{n}_i^*,\mathbf{n}_{R_h},\mathcal{T}(\cdot))=\left \|\mathbf{n}_i^* - {\mathcal{T}(\mathbf{n}_{R_h})} \right\|,
\end{equation}
where $\mathbf{R}$ is the estimated rotation matrix between RGB image and depth image. \par
Our chose to solve normal $\mathbf{n^*}$ instead of $p$ and $q$ is of advantages for two reasons. Firstly, $p$ and $q$ is unbounded, i.e. , the absolute value of $p$ or $q$ can be much larger than 1 while the elements in the normalized vector $\mathbf{n^*}$ vary among $\left[ -1,1\right]$. Secondly, in traditional SfS method, the relationship between $p$ and $q$ is not fully utilized. However, in the proposed method, $p$ and $q$ are solved together in the sense of $\mathbf{n}_i^*$, namely, 
\begin{equation}
\mathbf{n}_i^*=
\left[
\begin{matrix}
-\frac{{{p}_{i}}}{\sqrt{1+p _{i}^{2}+q _{i}^{2}}}, & -\frac{{{q }_{i}}}{\sqrt{1+p _{i}^{2}+q _{i}^{2}}}, & \frac{1}{\sqrt{1+p _{i}^{2}+q _{i}^{2}}}   
\end{matrix}
\right]^T,
\end{equation}
which is more suitable for the SH based shape from shading. 

\subsection{Optimization}
Since for $i$-th pixel a system of 3 equations can be set up for the 3 unknown variables $\mathbf{n_i^*}=\begin{bmatrix}
n_{i1}^* & n_{i2}^*  & n_{i3}^*
\end{bmatrix}^T$, trust-region-reflective algorithm is chosen to solve this well-posed nonlinear least square system. Compared with gradient descent methods, it does not need to compute actual partial derivatives of the function and runs faster generally. The optimization toolbox is integrated and provided in Matlab. The maximum number of iteration is set to be 1000, while in most cases the method will converge in less than 300 times of iteration.   

\subsection{Postprocessing}
Once the estimated unit normal $\mathbf{n_i^*}$ is obtained, we utilized the method on least squares surface reconstruction from gradients proposed by Harker \textit{et al} \cite{HarkerCVPR08,HarkerCVPR11} to solve the following system for $z_i^*$:
\begin{equation}
\left \{ \begin{matrix}
\frac{\partial z_i^*}{\partial x}=p_i\\ 
\frac{\partial z_i^*}{\partial y}=q_i
\end{matrix} \right.
\end{equation}
After obtaining the depth map $z_i^*(x,y)$, a point cloud $\mathbf{P}_c^*=\{ (x,y,z^*))\}$  can be formed for further registration.

\section{Point Cloud Registration}\label{5}
In this section, we will discuss how to estimate the pose $\mathcal{T(\cdot)}$ from $P_c$ to $P_c^*$, where $P_c$ is the point cloud from Kinect depth map $z_R$. This problem to estimate the pose can be viewed as estimating the scale factor $\mathbf{s}$, rotation matrix $\mathbf{R}$ and translation vector $\mathbf{t}$ by minimizing the correspondent point wise distance between estimated point clouds $P_c^*$ and $P_c$:
\begin{equation} \label{eq2}
f({{z}^{*}},s,\mathbf{R},\mathbf{t})=\underset{s,\mathbf{R},\mathbf{t}}{\mathop{\text{minimize}}}\,\ {{\left\| P_{c}^{*}-s\mathbf{R}{{P}_{c}}-\mathbf{t} \right\|}^{2}}
\end{equation}  \par
In one iteration, the estimated point cloud is fixed hence the problem is simplified as 3D point clouds registration. Iterative Closest Point (ICP) method is chosen to obtain an initial transform and set up the primary point-wise correspondences while RANSAC is chosen to reject the weak correspondences and get the refined pose $\mathcal{T(\cdot)}$. 
\par
\subsection{Pose Estimation}
The following form of RST transformation model is utilized in the RANSAC process:
\begin{equation}
{{T}_{\theta }}(\mathbf{y})=s\mathbf{Ry}+\mathbf{t}=\left[ \begin{matrix}
   {{a}_{11}} & {{a}_{12}} & {{a}_{13}}  \\
   {{a}_{21}} & {{a}_{22}} & {{a}_{23}}  \\
   {{a}_{31}} & {{a}_{32}} & {{a}_{33}}  \\
\end{matrix} \right]\mathbf{y}+\left[ \begin{matrix}
   {{t}_{1}}  \\
   {{t}_{2}}  \\
   {{t}_{3}}  \\
\end{matrix} \right],
\end{equation}
so that parameter vector is
\[\begin{aligned}
  \theta = &  [\ \begin{matrix}
   {{a}_{11}} & {{a}_{12}} & \begin{matrix}
   {{a}_{13}} & {{a}_{21}} & {{a}_{22}}  \\
\end{matrix}  \\
\end{matrix} \\ 
 & \begin{matrix}
   \quad {{a}_{23}} & {{a}_{31}} & \begin{matrix}
   \begin{matrix}
   {{a}_{32}} & {{a}_{33}} & {{t}_{1}}  \\
\end{matrix} & {{t}_{2}} & {{t}_{3}}  \\
\end{matrix}  \\
\end{matrix}{{]}^{T}}  
\end{aligned}\]

\par
Given $n$ point correspondences ${{\left\{ {{x}_{i}},{{y}_{i}} \right\}}_{i=1,..,n}}$ these transformation relationship can be grouped as following linear system
\begin{equation} \label{eq8}
\mathbf{b}=\left[ \begin{matrix}
   {{x}^{(1)}}  \\
   \vdots   \\
   \vdots   \\
   {{x}^{(n)}}  \\
\end{matrix} \right]=\left[ \begin{matrix}
   {{T}_{\theta }}({{y}^{(1)}})  \\
   \vdots   \\
   \vdots   \\
   {{T}_{\theta }}({{y}^{(n)}})  \\
\end{matrix} \right]=\mathbf{A\theta }
\end{equation}
Then the solution of Eq.~\eqref{eq8}, $\mathbf{\hat{\theta}}$ can be estimated through linear least square method.
\par 
Once got the optimal $\mathbf{\theta}$ and $\mathbf{R}$ when $CS$ between two point sets is the largest, the global rotation matrix $\mathbf{R}$ can be decomposed into the rotations along the three axis respectively:
\begin{equation}
\mathbf{R}   =s{{\mathbf{R}}_{x}}(\alpha ){{\mathbf{R}}_{y}}(\beta ){{\mathbf{R}}_{z}}(\gamma )
\end{equation}
where $\alpha$,$\beta$ and $\gamma$ are rotated angles, namely \textit{yaw, pitch, and roll} along x-axis, y-axis and z-axis respectively, $\mathbf{R}(\cdot)$ is the corresponding rotation matrix.\par 
Such problem can be easily formulated as a nonlinear least square estimation problem, which aims to obtain $\hat{\alpha}$,$\hat{\beta}$ and $\hat{\gamma}$ by minimizing 
\begin{equation} \label{eq9}
g(\alpha ,\beta ,\gamma )=\underset{\alpha ,\beta ,\gamma }{\mathop{\min }}\,{{\left\| {{\mathbf{R}}_{x}}(\alpha ){{\mathbf{R}}_{y}}(\beta ){{\mathbf{R}}_{z}}(\gamma )-\mathbf{R} \right\|}^{2}}
\end{equation}
\par 
Levenberg-Marquardt method is leveraged to optimize the Eq.~\eqref{eq9}. Compared to the trust-region-reflective algorithms, LM method is generally faster and hence will converge in several iterations in most cases.

\subsection{Point-wise Correspondences Set Up}
Theoretically, the estimated pose $\mathcal{T(\cdot)}$ can determine a pair of quadrilateral overlapping areas on depth maps $\mathbf{Z}^*$ and $\mathbf{Z}_R$ respectively. However, in practice depth maps are represented by point arrays. Therefore we set up the one-to-one mapping between points of overlap areas on the different depth maps in following three steps:
\begin{itemize}
\item[1] Map $P_c$ onto the space of $P_c^*$ by apply the transformation with parameters $\mathcal{T}(P_c)$ of estimated pose.
\item[2] For each point in $\{\mathcal{T}(P_c)\}$, find the 1st nearest neighbour in the point set $\{P_c^*\}$ and hence to form an initial correspondence.
\item[3] Accept the point wise correspondence based on the distance $d(P_c^*,\mathcal{T}(P_c))$ between $P_c^*$ and $\mathcal{T}(P_c)$. 
\end{itemize} \par
A naive modelling of acceptance criterion can be hard-counting, namely accepting the points pair if the correspondent distance $d(P_c^*,\mathcal{T}(P_c))$ is less than a threshold. In our method the threshold is  set to be 1.

\section{Shading-based Depth Refining}\label{6}
Given depth map may contain holes or noises.  Since normal $\mathbf{n}_{R_h}$ and depth $z_{R_h}$ of $h$-th pixel in the overlap area of given depth map are subjected to the shading model Eq.~\eqref{eq11} at $i$-th pixel on the RGB image, if the pose $\mathcal{T}(\cdot)$ is determined, we can refine the normal $\mathbf{n}_{R_h}$ in the a symmetric way of proposed SfS process, i.e. , 
\begin{equation}
\begin{aligned}
\min g(\mathbf{n}_{R_h}^*)& =g_1(\mathbf{n}_{R_h}^*,\mathcal{T}(\cdot))+{\lambda }_{g2}g_2(\mathbf{n}_i^*,\mathbf{n}_{R_h}^*,\mathcal{T}(\cdot)))\\
& +\lambda_{g3}({\mathbf{n}_{R_h}^*}^T\mathbf{n}_{R_h}^*-1).
 \end{aligned}
\end{equation}
The \textit{brightness constraint} at $k$-th normal on the given depth map can be formed as
\begin{equation}
{g_{\text{1}}}(\mathbf{n}_{R_h}^*)={{(\log S_ij-{(\mathbf{\hat{n}}_{R_h}^*})^T \mathbf{M}_j(\mathbf{\hat{n}}_{R_h}^*}))^{2}},
\end{equation}
where $\mathbf{\hat{n}}_{R_h}=\left [ \begin{matrix}
 \mathbf{R}^{-1}\mathbf{n}_{R_h}^*, & 1
\end{matrix} \right ]^T$. \par 
The shape prior term in the normal refinement is nothing but the symmetric version of Eq.~\eqref{eq10}:
\begin{equation}
g_{2}(\mathbf{n}_{R_h}^*)={{\left \|\mathbf{n}_{R_h}^*-{\mathcal{T}^{-1}(\mathbf{n_i^*})} \right\|}},
\end{equation}
where $\mathcal{T}^{-1}(\cdot)$ is the inverse transformation of the estimated pose $\mathcal{T}(\cdot)$, which maps $P_c^*$ onto $P_c$.
The refined depth at $h$-th pixel on the depth map $z_{R_h}^*$ can also be solved in the same manner of shape from surface gradients method.

\section{Experiments}\label{7}
In this section, we will validate the proposed method on both synthesized data and real RGB-D data. Both qualitative and quantitative analysis are conducted not only to show the results of our method, but also to explore the reason why our method may work and how the method perform under different pose and different wideness of overlap area.
\subsection{Synthetic Study}
\begin{figure}
\centering
       \includegraphics[width=0.5\textwidth]{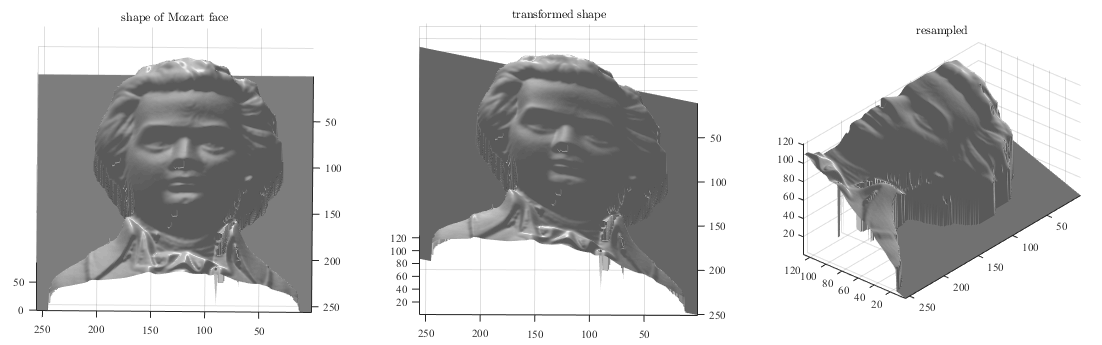}
      \caption{The illustration of the data synthesizing process. The frontal depth map of Mozart's face(\textit{left figure}) is rotated (\textit{middle figure}) and clipped (\textit{right figure}). Afterwards, the log-RGB image can be generated through Eq.~\eqref{eq11}.}
      \label{fig3}
\end{figure}

Two quantitative experiments are designed and conducted to explore how and to what extent the shape prior information can be utilized in the shape from shading process as well as to validate the performance of our methods on the $32\times 32$ synthesized pair of RGB image and rotated depth map of Mozart's face. The synthesis data can be obtained in the following steps, shown in Figure~\ref{fig3}: \\
a. Given certain SH coefficients and a frontal depth map of Mozart's face, the log-RGB image can be obtained through Eq.~\eqref{eq11}.\\
b. Given certain rotation angle $\alpha,\beta,\gamma$ along x-,y- and z-axis respectively, the rotated depth map can be generated.\\
c. The wideness of the baseline is determined by the grid we use to sample the rotated depth map.\\

\begin{figure*}[!htp]
\centering
       \includegraphics[width=\textwidth]{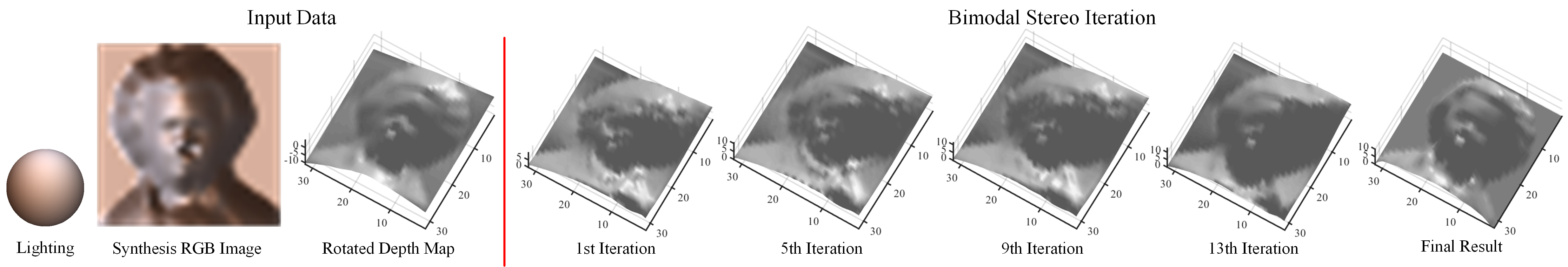}
      \caption{Experiment results on the synthesized images of Mozart face, which shows during the process the estimated depth $z^*$ is obviously improved, especially the right upper part of the shape.}
      \label{fig4}
\end{figure*}

Figure~\ref{fig4} shows the experiment conducting on the synthesis RGB image under given lighting conditions and depth map of Mozart's face from another view. The depth map is obtained by rotating the ground truth depth for 20 degrees along y-axis. 
The loop terminated after 16 iterations. During the process the estimated depth $z^*$ is obviously improved, especially the right upper part of the shape. The estimated pose paramters are listed in Table~\ref{table1}.

Besides, we conducted two quantitative experiments to further study the capability of the proposed method. One is to explore the reason why shape prior can improve the reconstruction result of the proposed SfS method, the other is to discuss the performance under different rotations and wideness of baseline.    
\begin{figure*}[!htp]
\centering
       \includegraphics[width=\textwidth]{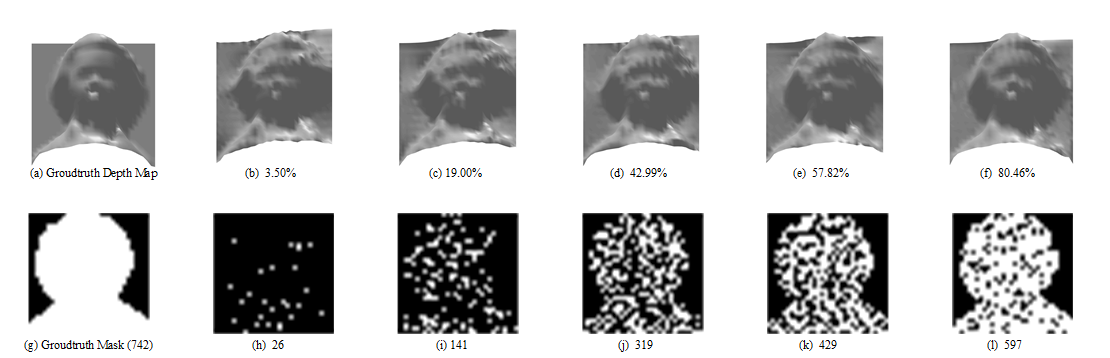}
      \caption{This figure shows how the reconstructed result improves with respect to the increase of the shape prior. Here we use a synthetic $32\times 32$ depth map of Mozart's face as shown in (a), the binary mask of which is shown in (g). The rest figures of the upper row, (b)-(f), shows the reconstructed surfaces by SfS while the lower, (h)-(l) , shows the location and amount of prior points are used in this process.}
      \label{fig5}
\end{figure*}

\begin{figure*}[!htp]
\centering
       \includegraphics[width=\textwidth]{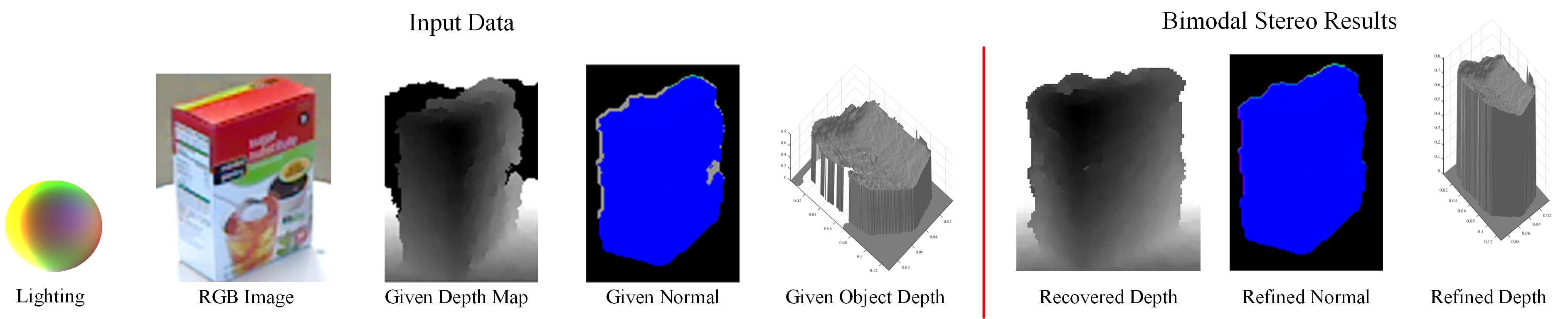}
      \caption{Experiment results on the RGB image and depth map of a foodbox proposed by Lai\cite{LaiICRA11}. The given normal and depth are shown improved after the refinement. Moreover a plausible depth map of the RGB image is obtained.}
      \label{fig7}
\end{figure*}

\subsubsection{Experiments on Shape Prior} \label{prior}
In the first experiment, we will validate the underlying assumption of our method, i.e. by posing the point-wise shape prior term in the original brightness equation Eq.~\eqref{eq12}, the SfS reconstruction can be improved.\par
The experiment is conducted by controlling the percentage $P_{er}$ of the shape prior provided. $P_{er}=0$ when no prior is offered while $P_{er}=1$ means every point is given the prior information.

The result of this experiment can be shown in Figure~\ref{fig5}. According to the results, recovered shape is improved and approaching the ground truth when $P_{er}$ is increased. In this sense we can conclude that the SfS reconstruction result of certain image area can be improved by given shape prior of this area.


\subsubsection{Experiments on Different Rotation and Baseline}
The next experiment is to discover the method performance in ''Bimodal'' settings, i.e. the depth map and RGB images are synthesized from different views. We'd like to simulate the situation where the depth camera and RGB camera shares partial overlap views and study the relationship between estimation accuracy and parameters, i.e. rotation and baseline, of given depth map. The term 'baseline' from the Stereopsis to describe the distance of two optical centers, which describes the wideness of overlap area.\par
\begin{table} \small
\centering
\begin{tabular}{c  c  c  c  c }
\toprule
   & s & $\alpha$ & $\beta$ & $\gamma$ \\
\midrule
Esimated & 1.0000 & 0.03216$^\circ$ & 18.7013$^\circ$ & 0.01812$^\circ$\\ 
Groundtruth & 1.0000 & 0$^\circ$ & 20$^\circ$ & 0$^\circ$\\
\bottomrule
\end{tabular}
\caption{Estimated pose paramters of Figure~\ref{fig4}}
\label{table1}
\end{table} 
\begin{table} \tiny
\centering
\begin{tabular}{c  c  c   c  c  c  c  }
\toprule
$R^*$ / $P_w$  & 0.125 & 0.25 & 0.375 & 0.5 &0.75 & 1 \\
\midrule
20 & 0.022806 & 0.015999 & 0.006442 & 0.006005 &0.007384 &0.008664\\ 
40 & 0.022903 & 0.011479 & 0.006860 & 0.005768& 0.005786 & 0.006653\\
60 & 0.022379 & 0.018440 & 0.012250 & 0.007660 & 0.004729& 0.004509 \\
80 & 0.010623 & 0.008897 & 0.008637 & 0.005853 &  0.002500&0.001922 \\
90 & 0.007870 &1.631535 & 0.006505 & 0.004082 & 0.000622& 0.000649\\
\bottomrule
\end{tabular}
\caption{Error of estimated rotation under different rotate angles and different wideness of overlap area.}
\label{table2}
\end{table}
\begin{table} \small
\centering
\begin{tabular}{c  c  c  c  c }
\toprule
   & s & $\alpha$ & $\beta$ & $\gamma$ \\
\midrule
Esimated Pose& 0.9972 & -0.4813$^\circ$ & 17.3492$^\circ$ & 9.3220$^\circ$\\
\bottomrule
\end{tabular}
\caption{Estimated pose paramters of Figure~\ref{fig7}}
\label{table3}
\end{table} 
In the experiment, the rotation angle along y-axis $\beta$ is chosen from $\{20,40,60,80,90\}$ and the wideness of overlap area $P_{w}$, i.e. the percentage of the total area, is chosen from $\{0.125,0.25,0.375,0.5,0.75,1\}$. The normalized difference of estimated rotation $R^*$ and ground truth $R$ is computed in the following equation:
\begin{equation}
err=\left \| \frac{\textbf{R}^*}{\left \|\textbf{ R}^* \right \|}- \frac{\textbf{R}_g}{\left \|\textbf{ R}_g \right \|}\right \|
\end{equation} \par
\begin{figure}
\centering
       \includegraphics[width=0.45\textwidth]{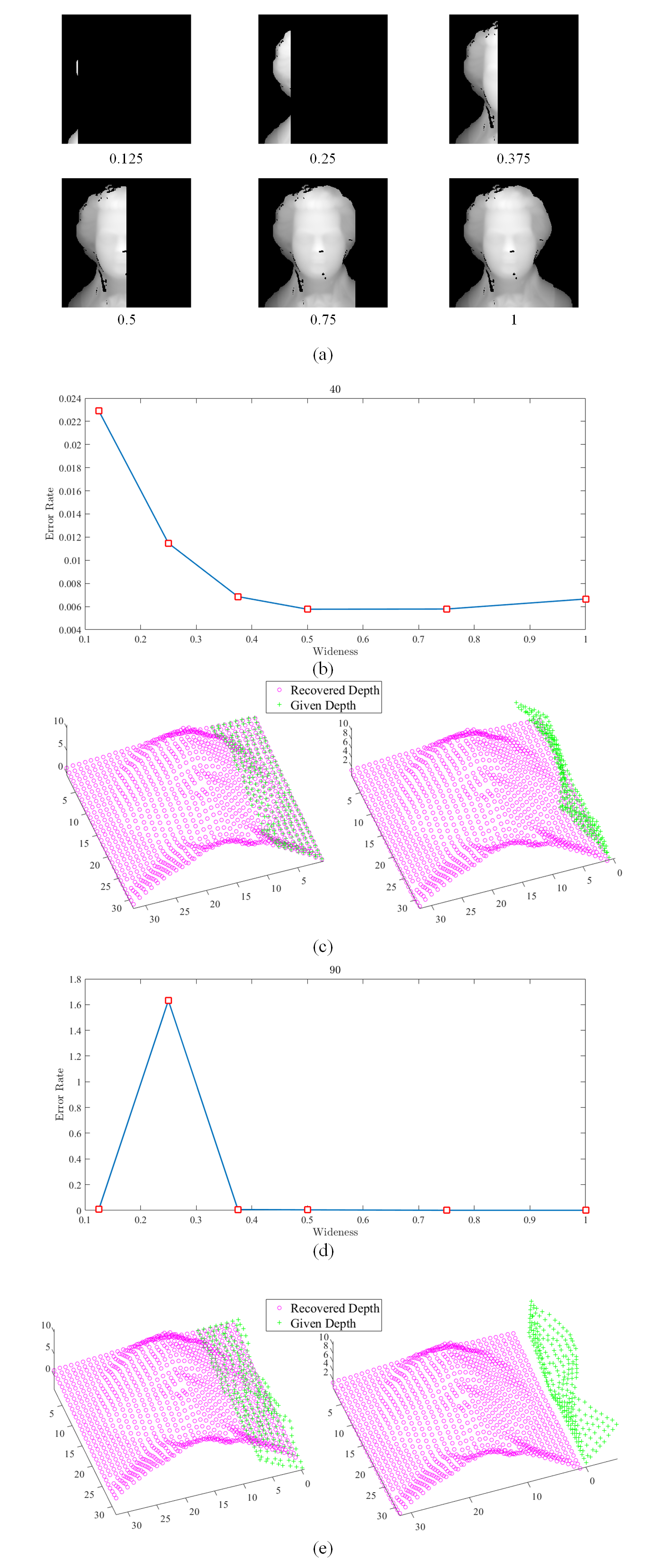}
      \caption{The illustration of experiments on the the relationship between estimation accuracy and parameters, i.e. rotation and baseline , of given depth map. Fig.(a) shows the synthesized depth images containing overlap area of certain wideness $P_w$ . In (b) and (c) we show the error variation at 40$^{\circ}$ when wideness changes as well as the visualization (right side in (c)) and registration (left side in (c)) of estimated shape and given point cloud. The error is decreasing when the overlap area becomes larger. In (d), the failure case is illustrated when $\beta$ reaches 90 and $P_w$ is 0.25, point clouds before registration are visualized in (e).}
      \label{fig6}
\end{figure}

The numerical results are shown in Table~\ref{table2} while in Figure~\ref{fig6} the illustration of wideness of the overlap area is shown in (a). In (b) and (c) we show the error variation at 40$^{\circ}$ when wideness changes as well as the visualization (right side in (c)) and registration (left side in (c)) of estimated shape and given point cloud. The error is decreasing when the overlap area becomes larger. However in (d) of Figure~\ref{fig6}, the failure case appears when $\beta$ reaches 90 and $P_w$ is 0.25, which is visualized in (e).\par
As shown, since the intrinsics of the depth camera are taken as known parameters and hence can obtain the point cloud easily, the method is robust in most rotations and wideness of overlap area.

\subsection{Real Data Study}
The proposed Bimodal stereo method is validated on RGB-D Object Dataset proposed by Lai\textit{ et al} \cite{LaiICRA11}. This dataset contains over 300 instances in 51 categories, where the dataset of each instance is composed of cropped RGBD images of the object from different views. The experiments are conducted on the data of \textit{foodbox}, \textit{mug} and  \textit{lemon}, etc. \par
The reconstruction and refinement on the instance \textit{foodbox} is displayed in Figure~\ref{fig7} while the estimated pose parameters are shown in Table~\ref{table3}.

In this case, we chose \textit{foodbox-1-1-5} as RGB input and \textit{foodbox-1-1-18} as depth input, the pose between which could be observed obviously. Besides, the depth map contains small holes on the surface of foodbox, which shouldn't have been there. \par 
The improvement of the given normal and depth are shown in the results part of Figure~\ref{fig7}. Moreover, after 19 times of the iteration, a plausible depth map of \textit{foodbox-1-1-5} is obtained. Such results can validate the usefulness of the proposed method.

\section{Discussion and conclusion}
\textbf{Limitations.} Although the capability of our method has been proved robust in the previous experiments, it should be pointed out that like other Shape from shading method or shading based shape refinement solved by non-linear least square method, the most limited part is the time and space costs of the optimization . For example, a single iteration on $32\times32$ images may take 10 minutes. Thus for an experiment on images of such size can take several hours. \par
\textbf{Conclusion.} In this paper, a new problem called "Bimodal Stereo" is established, which literately stereopsis from two different views of different modality, aims to recover the global shape and estimating rigid transformation between RGB camera and depth camera simultaneously. To solve this problem, we proposed an iterative computational framework which regularizes SfS process with shape prior from depth map while estimates the pose parameters between recovered shape and rotated depth map. Quantitative experiments showed that through our method, both SfS results and depth map can be refined as well as the pose parameters are robustly estimated.


\bibliographystyle{spmpsci}      
\bibliography{test}   

\end{document}